\pdfoutput=1

\documentclass[11pt, table, dvipsnames]{article}

\usepackage[preprint]{acl}

\usepackage{times}
\usepackage{latexsym}

\usepackage[T1]{fontenc}

\usepackage[utf8]{inputenc}

\usepackage{microtype}

\usepackage{inconsolata}

\usepackage{graphicx}

\usepackage{booktabs}       
\usepackage{amsfonts}       
\usepackage{nicefrac}       
\usepackage{microtype}      

\usepackage{hyperref}       
\usepackage{cleveref}
\usepackage{wrapfig}
\usepackage{multirow}
\usepackage{comment}
\usepackage{makecell}
\usepackage{caption}
\usepackage{subcaption}
\usepackage{float}
\usepackage{placeins}

\title{\textsf{RESPONSE}: Benchmarking the Ability of Language Models to Undertake Commonsense Reasoning in Crisis Situation}

\author{A\"issatou Diallo$^1$\thanks{ Corresponding author: \texttt{a.diallo@ucl.ac.uk}\\}, Antonis Bikakis$^2$, Luke Dickens$^2$, Anthony Hunter$^1$, Rob Miller$^2$ \\
$^1$Department of Computer Science \\
$^2$Department of Information Studies     \\
University College London, United Kingdom \\}

\begin{document}
\maketitle


\begin{abstract}
An interesting class of commonsense reasoning problems arises when people are faced with natural disasters. To investigate this topic, we present \textsf{RESPONSE}, a human-curated dataset containing 1789 annotated instances featuring 6037 sets of questions designed to assess LLMs' commonsense reasoning in disaster situations across different time frames. The dataset includes problem descriptions, missing resources, time-sensitive solutions, and their justifications, with a subset validated by environmental engineers. Through both automatic metrics and human evaluation, we compare LLM-generated recommendations against human responses. Our findings show that even state-of-the-art models like GPT-4 achieve only 37\% human-evaluated correctness for immediate response actions, highlighting significant room for improvement in LLMs' ability for commonsense reasoning in crises. 
\end{abstract}

\section{Introduction}


As LLMs are being developed to handle different kinds of reasoning, there are questions about their ability in diverse commonsense reasoning situations. An interesting class of commonsense reasoning scenarios arises when people are faced with natural disasters. For this, it would be desirable for LLMs to be able to identify problems and solutions that are based on clear steps of commonsense reasoning, starting from the particular circumstances of the disaster scenario. 
\begin{figure}[ht]
  \centering
    \includegraphics[width=0.42\textwidth]{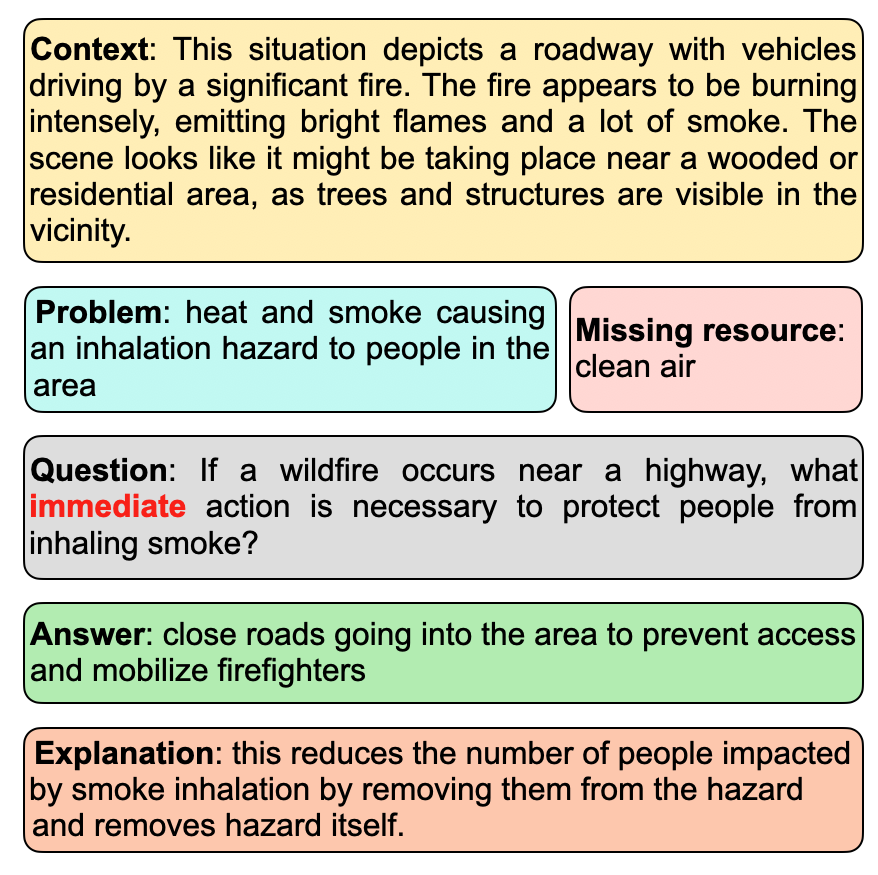}
  \caption{A sample from \textsf{RESPONSE} for the "immediate" time frame. The type of incident and location are (\textit{wildfire}, \textit{highway}). }
  \label{fig:sample}
\end{figure}

While it may seem concerning to rely on LLMs for disaster management advice, the reality is that people are already turning to these systems for guidance in various situations. At the same time, LLMs are becoming indispensable tools, serving in several roles such as dialogue agents, data analysts, and decision support. One critical area where LLMs could make a significant impact is in disaster analysis and response, particularly during preparation and recovery phases when immediate danger isn't present. To exploit this potential effectively, we must evaluate their alignment with human commonsense in real-world disaster scenarios. If aligned, LLMs could provide valuable assistance to both individuals and agencies - offering support when traditional aid is limited, and helping agencies with scenario planning to develop more effective response strategies.

The range of emergencies is vast, from localized flooding to widespread wildfires. Their consequences, both immediate and long-term, can be devastating. To ensure LLMs are truly helpful, we must test them against a variety of scenarios, incorporating factors like time constraints, resource shortages, and specific geographic contexts. This will help us determine if LLMs can reason beyond regurgitating generic best practices.
More specifically, a proper evaluation should consider all aspects that make a remedial solution to an incident functional. When prompted with a incident-related question with a specific time setting, is the response \textit{sufficient and appropriate} for the situation? Is the recommended course of action \textit{useful}? Does the response contain all the information needed to act? And finally, \textit{would a human follow this solution?} These are the questions explored in this work. 

We propose a comprehensive dataset specifically designed to test the quality of LLM responses in disaster situations. This allows us to evaluate how LLMs interpret information, plan courses of action, and ultimately, provide good quality advice and recommendations during critical incidents. 

Building on existing work on incident and natural disaster detection \cite{weber2020detecting,weber2022incidents1m}, we have compiled a natural language dataset covering a variety of natural disasters, contextualized in different locations and time frames.
Specifically, the dataset consists of natural language questions, each addressing a specific incident-related problem that humans might face. We collected human annotations identifying the main problem and the missing resource as short sentences. Additionally, the annotators were asked to provide time-dependent solutions to the identified problem and explanations for their reasoning. Then, we automatically generated the questions using the identified problem and missing resource. These are contextualized in a specific temporal time frame and location. Furthermore, each set of questions is associated with the solutions and explanations collected. An example of a question from \textsf{RESPONSE} is: \textit{"In the event of an ice storm leaving vehicles stranded in a \underline{snowfield}, what is the \underline{first step} to take to assist those trapped?"}, and the associated gold answer is \textit{"plow snow enough to get a pathway for cars"}. The underlined text represents the location and time frame, respectively.

\textit{We made the deliberate choice to collect annotations from general annotators to test how LLMs align with human commonsense in disaster management}. To validate the quality of these annotations, we consulted environmental engineers who evaluated a sample of the collected responses. This dataset is not intended to: (i) \textbf{Replace human expertise}: the dataset aims to assess LLM performance in comparison to human judgments, while acknowledging the critical role of human experts in real-world disaster situations. (ii) \textbf{Provide definitive solutions}: while the dataset includes human-annotated solutions that have been partially validated by domain experts, these are not necessarily the only possible responses. The goal is to evaluate how well LLMs can align with human reasoning when providing relevant advice.

Next, we used the resulting dataset as a benchmark to evaluate thecommonsense recommendations of two of the best performing proprietary LLMs. We evaluated the responses using common evaluation metrics as well as human evaluation. Our findings show that, (1) LLMs generate solutions and explanations for incident related questions that are judged better than humans only in 37\% of cases for the immediate time frame, and this value increases to 58\% as the time horizon increases; (2) none of the commonly used evaluation metrics can properly discriminate the solutions.

\begin{figure*}[t]
    \centering
    \includegraphics[width=\textwidth]{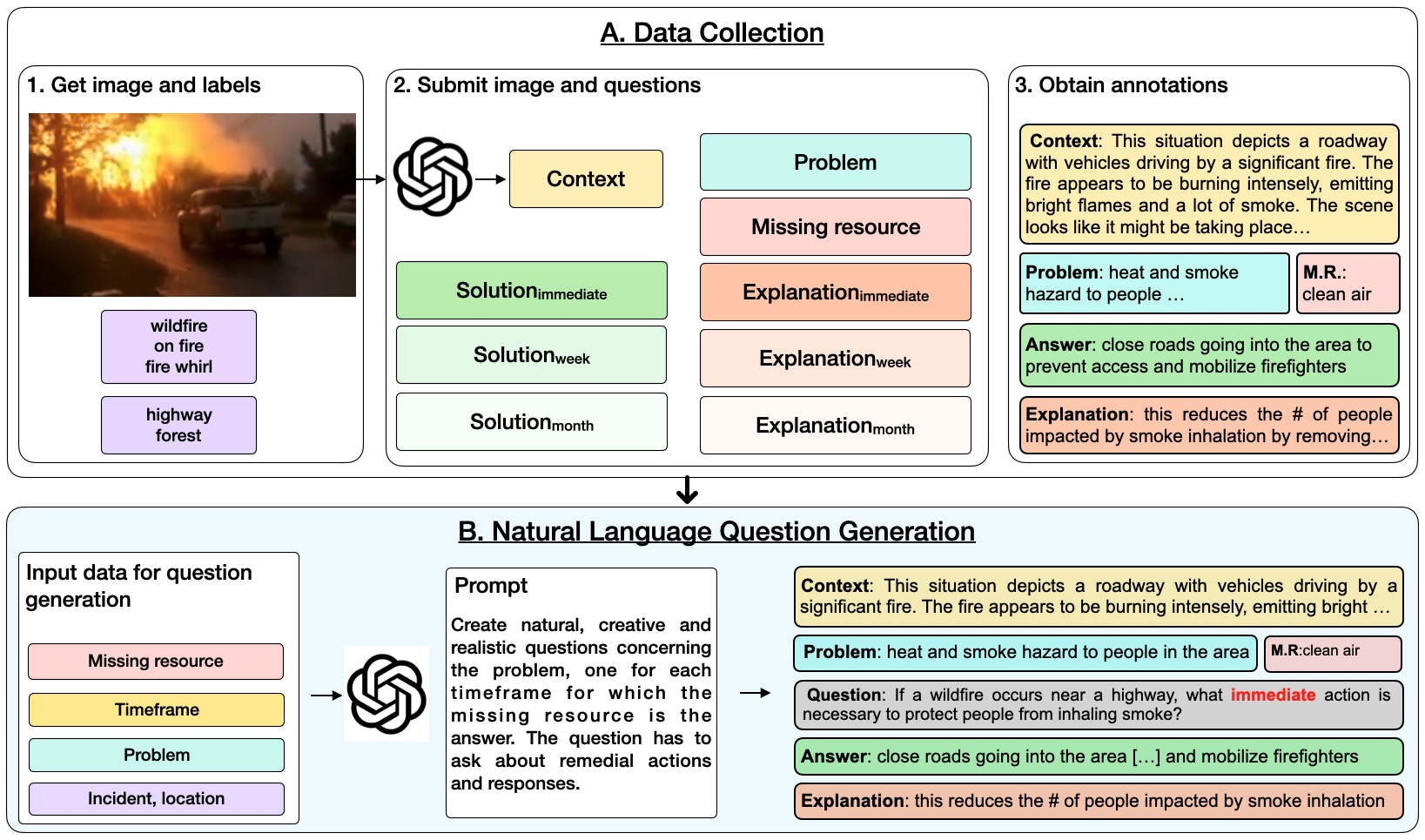}
    \caption{Data-generating process for the \textsf{RESPONSE} dataset. Top: Image and labels annotated with problem, missing resource (MR), solutions, and explanations. Bottom: Natural language question generation based on problem, missing resource, and time frame.}
    \label{fig:pipeline}
\end{figure*}

\textbf{Contributions} In summary, our contributions are: (1) we created \textsf{RESPONSE}, a question-answer (QA) dataset focusing on commonsense reasoning in disaster management, with annotations validated by two environmental engineers to ensure response quality. To the best of our knowledge, this is the first dataset of natural disaster management requiring commonsense reasoning to answer incident related questions with short natural language explanations. (2) We perform experiments on state-of-the-art LLMs to benchmark the commonsense reasoning of LLMs in disaster situations. (3) We analyze the alignment between LLM-generated disaster management strategies and those proposed by humans.

\section{Related Work}
This work is in the same line of thinking as existing
benchmarks evaluating how LLMs align with humans thinking such as \cite{santurkar2023whose}, which probes the alignment of LM opinions with US demographic groups or \cite{jin2022make}, which explores LLMs’ understanding and alignment of human moral reasoning.

\paragraph{Skill evaluation for LLMs} This work can be seen as part of the recent literature aimed at evaluating the reasoning capabilities and performance of LLMs. Specifically, there are various studies investigating LLMs capabilities in domains such as medicine \cite{nori2023capabilities,singhal2023large}, education \cite{baidoo2023education,rudolph2023chatgpt}, law \cite{katz2024gpt}, social science\cite{ziems2024can}. To the best of our knowledge, this is the first work aiming at evaluating LLMs situational awareness in incidents and disaster management.

\paragraph{LLMs and Human Behaviour}  There is a growing literature investigating LLMs acting as simulators of human agents \cite{argyle2023out,horton2023large,park2023generative}. These models encode a wide range of human behavior from their training data \cite{bommasani2021opportunities}. When prompted with a narrowly defined context, LLMs can be used to generate believable behavior. Existing studies have replicated existing social science studies \cite{horton2023large}, simulated political surveys \cite{sorensen2022information}, and generated synthetic data \cite{hamalainen2023evaluating} using this approach. LLMs have also been used to create interactive human behavior for users to engage with, such as in interactive fiction \cite{freiknecht2020procedural} and text adventure games \cite{callison2022dungeons}. Our study aims to supplement this growing literature by assessing how LLMs simulate and understand human behaviour in contexts that require situational awareness, disaster management knowledge and commonsense to react to incidents. 

\paragraph{LLMs for Disaster Management} 
ChatGPT's capabilities in natural disaster management were explored in \cite{xue2023application}. The study builds upon the potential of LLMs to streamline disaster response across pre-disaster, during-disaster, and post-disaster stages. Their work highlights the limitations of current methods in real-time data handling and achieving expert-level accuracy in complex scenarios. Similarly, this work \cite{goecks2023disasterresponsegpt} propose the DisasterResponseGPT framework, leveraging LLMs for rapid action plan generation. However, both approaches face limitations inherent in general-purpose LLMs like ChatGPT. First, the lack of specialized training or domain-specific knowledge might lead to doubt in ChatGPT's ability to provide sufficiently accurate human commonsense answers concerning complex disaster prevention and mitigation strategies. Secondly, the challenge of explaining and verifying generated responses creates difficulties for users in assessing their reliability and trustworthiness. Our dataset specifically addresses these two key shortcomings, in order to fully assess the true potential of LLMs in disaster management with respect to commonsense reasoning. 

\section{The RESPONSE Dataset}

The objective of this study is to present a high-quality and challenging dataset of QA sets with the corresponding explanations. We developed an extensive data collection pipeline using the crowdsourcing platform CloudConnect Research.

\paragraph{Motivation}
Our dataset focuses on evaluating the ability of LLMs to answer questions about disaster using commonsense, rather than relying on general expert-defined rules. This approach is motivated by the real-world challenges faced by decision-makers in disasters, where they often need to make informed judgments based on available information and general understanding. By focusing on missing resources, we isolate a critical factor that can significantly impact decision-making. Temporal horizons are included to simulate the dynamic nature of disaster response, where different actions may be appropriate at different stages.

Explanations are required to assess LLMs' understanding of the problem, their ability to identify missing resources, and their justification for proposed solutions.  
While expert annotations could provide valuable insights, our reliance on human raters aligns with the real-world scenario where decision-makers often rely on their own judgment and general knowledge for decision-making. Moreover, it allows us to investigate the role of commonsense.

\paragraph{Closer to a probe than a benchmark} While the potential for incorrect or misleading advice from LLMs is a serious concern, our dataset serves as an analytical tool to understand model behavior in disaster response scenarios. Rather than serving as a benchmark for optimization, our dataset helps researchers identify where models succeed or fail in using commonsense when providing practical disaster management advice, particularly in understanding temporal dependencies and resource constraints.

\begin{table*}[h]
\centering
\resizebox{0.95\textwidth}{!}{%
\begin{tabular}{llccc}
\hline
\multicolumn{1}{c}{\textbf{Setting}} & \multicolumn{1}{c}{\textbf{Solution}}                                             & \textbf{Suff.} & \textbf{Usef.} & \textbf{Correct.} \\ \hline
\rowcolor{lightgray}
\multicolumn{5}{l}{\makecell[l]{\textit{Immediately after a flood near a river disrupts transportation, what action should be prioritized to}\textit{address the issue?}}}                \\

\multirow{2}{*}{Gold}                & Answer: announce the news of the flooding                                         & \multirow{2}{*}{-}   & \multirow{2}{*}{-}  & \multirow{2}{*}{-}   \\
                                     & Explanation: this will keep the residents of the area informed and save lives             &                      &                     &                      \\
\multirow{2}{*}{GPT-4}               & Answer: seek higher ground immediately, avoid flood waters.                       & \multirow{2}{*}{-1}  & \multirow{2}{*}{-1} & \multirow{2}{*}{-1}  \\
                                     & Explanation: floodwaters can rise quickly, posing immediate danger.                       &                      &                     &                      \\
\multirow{2}{*}{Claude}              & Answer: assess road safety and establish alternate routes.                        & \multirow{2}{*}{-}   & \multirow{2}{*}{-}  & \multirow{2}{*}{-}   \\
                                     & Explanation: ensuring public safety and maintaining essential transportation are crucial. &                      &                     &                      \\ \hline
                                     \rowcolor{lightgray}
\multicolumn{5}{l}{\textit{What immediate steps should you take to protect your flat home from becoming flooded near a river?}}                                                              \\
 
\multirow{2}{*}{Gold}                & Answer: use sand bags to slow down the flow of water.                             & \multirow{2}{*}{-}   & \multirow{2}{*}{-}  & \multirow{2}{*}{-}   \\
                                     & Explanation: it's a quick solution and nonpermanent                                       &                      &                     &                      \\
\multirow{2}{*}{GPT-4}               & Answer: avoid contact, report to authorities immediately.                         & \multirow{2}{*}{-1}  & \multirow{2}{*}{0}  & \multirow{2}{*}{-1}  \\
                                     & Explanation: quick reporting helps manage and mitigate risks.                             &                      &                     &                      \\
\multirow{2}{*}{Claude}              & Answer: sandbag or barricade entrances and low-lying areas.                       & \multirow{2}{*}{-}   & \multirow{2}{*}{-}  & \multirow{2}{*}{-}   \\
                                     & Explanation: prevents water from entering and diverts flow away from home.                &                      &                     &                      \\ \hline
                                     \rowcolor{lightgray}
\multicolumn{5}{l}{\textit{What should be done immediately if a forest catches fire to mitigate the impact on life and housing?}}                                                            \\
 
\multirow{2}{*}{Gold}                & Answer: relocate those with houses nearby to shelters.                            & \multirow{2}{*}{-}   & \multirow{2}{*}{-}  & \multirow{2}{*}{-}   \\
                                     & Explanation: this would help save lives in the affected area                              &                      &                     &                      \\
\multirow{2}{*}{GPT-4}               & Answer: evacuate area, alert authorities, assess immediate dangers.               & \multirow{2}{*}{1}   & \multirow{2}{*}{1}  & \multirow{2}{*}{1}   \\
                                     & Explanation: prioritize safety, coordinate emergency response efforts.                    &                      &                     &                      \\
                                     & Explanation: prioritize human lives, then property. Alert professionals to contain fire.  &                      &                     &                      \\ \hline
\end{tabular}}
\caption{Human evaluation for the "immediate" timeframe. (-1, 0, 1) means (\textit{lose}, \textit{tie}, \textit{win}) respectively.}
\label{tab:output_imm_part}
\end{table*}

\paragraph{Task Formulation} 
The dataset need to provide insights into the strengths and weaknesses of LLMs in navigating these complex situations, allowing us to measure their suitability for supporting human decision-making in incident and disaster response. 
Like the example in \Cref{fig:sample}, \textsf{RESPONSE} consists of N sets each containing a problem $p_i$, a missing resource $m_i$, a question $q^t_i$ and the corresponding $a^t_i$. $t$ represents the time frame for a specific question ["immediate", "after a week", "after a month"]. Previous works investigated reasoning through multiple choice \cite{lu2022learn}, retrieval \cite{asai2019learning}, and partial generation \cite{dalvi2021explaining}, however their task formulation has often the drawback of not revealing the reasoning process. For this reason, we opt for a generative task. The QA format evaluates the ability of LLMs to understand problems, identify missing resources, propose solutions, and justify their reasoning.

\paragraph{Preprocessing} We began with sampling 10000 data instances from Incidents 1M \cite{weber2022incidents1m} which ensures that the initial set of images are related to incidents and disasters. Incidents 1M is a large-scale dataset comprising over one million labeled images, designed for the task recognizing and categorizing incidents within visual scenes. It contains 43 incidents (e.g. landslide, flooding, earthquake, etc) and 49 location categories (e.g. building outdoor, highway, mountain). From this set of incidents, we exclude all the traffic related ones as well as the incidents were human survival is unlikely such as nuclear explosion, or again incidents that the common person is unlikely to observe such as "volcanic eruption". Then, we downloaded the images corresponding to this subset. Due to the inherent noise in the images from such a large dataset, we manually select the most expressive images. 

\paragraph{Data Collection}
We recruited college-educated annotators from English-speaking regions through CloudConnect Research, requiring basic familiarity with disaster scenarios. Using a structured interface, annotators viewed incident images and identified: (1) the main problem someone might face, (2) key missing resources, and (3) time-sensitive solutions with explanations for three horizons (immediate, week, month). Solutions were limited to 15 words. For each scenario, we collected annotations from two different annotators and had environmental engineers validate a sample for quality assurance. The interface facilitated structured responses by providing separate fields for each component, as shown in \Cref{fig:pipeline}.

\paragraph{Natural Language Question Generation} After manually verifying the quality of the annotations, we prompted GPT-4 to create natural questions for the collected samples in a 1-shot manner. We provided a handcrafted example of the input data and output questions. Specifically, we first prompted the LLM to provide all the possible combinations of place and incident labels. Then, we asked GPT-4 to create one question for each tuple \textit{(incident, place)}, from the original Incidents 1M dataset classes. We specified in the prompt, that the questions should mention the main problem, place and incident label and be related to the missing resource. Solutions and explanations were not included in the prompt. Additionally, each tuple \textit{(incident, place)} should correspond to three natural questions, one for each time frame.

\paragraph{Entry Contents} The \textsf{RESPONSE} dataset contains 895 unique samples, with two sets of annotations each (1789 total) from different annotators to capture diverse perspectives. Each sample consists of: (1) a \textit{problem} description from the annotator's perspective, (2) an identified \textit{missing resource}, (3) three time-dependent solutions (immediate, week, month) addressing the resource shortage, (4) explanations justifying each solution, and (5) a GPT-4 generated \textit{context} describing the original image from Incidents 1M. This structure allows \textsf{RESPONSE} to function as either an image-text or text-only dataset. \Cref{tab:stats} presents dataset statistics and \Cref{fig:pie_chart} shows the distribution of missing resources.

\section{Experiments}

\subsection{Experimental Setup}

\paragraph{Standard Question Answering} We evaluated LLMs' ability to provide remedial solutions to incident-related questions. To ensure fair comparison with human annotations, we constrained LLM responses to 15 words, matching the limit given to human annotators in \textsf{RESPONSE}.
\paragraph{Question Answering and Explanation} We extended the evaluation to include both solutions and their justifications, as clear reasoning is crucial for disaster response recommendations. Models were prompted to provide both an answer and explanation, each limited to 15 words.
\paragraph{Models Evaluated}
We tested two state-of-the-art LLMs: GPT-4 (\texttt{gpt-4-turbo}) \cite{achiam2023gpt} and Claude-3 Opus (\texttt{claude-3-opus}) \cite{claude3}. Both models were queried with temperature set to 0 for maximum consistency.

\paragraph{Baselines} 
Our empirical investigation focuses on some of the most recent LLMs. We test on the latest GPT-4 \cite{achiam2023gpt} (\texttt{gpt-4-turbo}) and the latest Claude-3 Sonnet (\texttt{claude-3-sonnet})\cite{claude3}. Both models were queried with temperature set to 0 for maximum consistency. 

\begin{table*}[h]
\resizebox{\textwidth}{!}{%
\begin{tabular}{@{}lllllllllll@{}}
\toprule
\multirow{2}{*}{\textbf{Timeframe}} & \multirow{2}{*}{\textbf{Model}} & \multicolumn{3}{c}{\textbf{Answer}}            & \multicolumn{3}{c}{\textbf{Explanation}}       & \multicolumn{3}{c}{\textbf{Answer + Explanation}} \\ \cmidrule(l){3-5}  \cmidrule(l){6-8} \cmidrule(l){9-11}
                                    &                                 & BLEURT         & BertS         & BLEU          & BLEURT         & BertS         & BLEU          & BLEURT          & BertS          & BLEU           \\ \midrule
\multirow{2}{*}{\textbf{Immediate}} & \textit{GPT-4}                  & \textbf{-0.66} & 0.59          & 0.44          & -0.85          & 0.55          & 0.04          & \textbf{-0.62}  & 0.57           & 0.42           \\
                                    & \textit{Claude-3}               & -0.69          & \textbf{0.60} & \textbf{0.68} & \textbf{-0.83} & \textbf{0.56} & \textbf{0.09} & -0.63           & \textbf{0.58}  & \textbf{0.51}  \\ \midrule
\multirow{2}{*}{\textbf{Week}}      & \textit{GPT-4}                  & \textbf{-0.68} & 0.59          & 0.17          & \textbf{-0.76} & \textbf{0.56} & 0.03          & \textbf{-0.61}  & 0.56           & 0.12           \\
                                    & \textit{Claude-3}               & -0.75          & 0.59          & 0.17          & -0.83          & 0.55          & \textbf{0.04} & -0.64           & \textbf{0.57}  & \textbf{0.13}  \\ \midrule
\multirow{2}{*}{\textbf{Month}}     & \textit{GPT-4}                  & \textbf{-0.70} & 0.60          & 0.29          & \textbf{-0.74} & 0.56          & 0.06          & \textbf{-0.61}  & 0.56           & 0.19           \\
                                    & \textit{Claude-3}               & -0.76          & 0.60          & \textbf{0.37} & -0.76          & 0.56          & 0.06          & -0.62           & \textbf{0.57}  & \textbf{0.28}  \\ \bottomrule
\end{tabular}}
\caption{Performance of GPT-4 and Claude-3 based on automatic metrics. The best scores are in bold.}
\label{tab: auto_eval_results}
\end{table*}

\subsection{Automatic Evaluation Metrics}

The still developing area of text generation has measures and proxies for correctness of text sequences based on similarity. For this reason, we used automatic evaluation metrics\footnote{\scriptsize{\url{https://huggingface.co/evaluate-metric}}} that measure surface similarity of the predictions to the references. 
\textbf{BLEU} \cite{papineni2002bleu} (Papineni et al., 2002) is one of the most known evaluation metrics for text generation. It computes the geometric average of the precision over output text’s n-grams. We use SacreBLEU \cite{post-2018-call} that produces comparable and reproducible BLEU scores.
\textbf{BertScore} \cite{zhang2019bertscore} uses BERT \cite{devlin-etal-2019-bert} embeddings and matches words in the output and reference sentences by cosine similarity. It then computes precision, recall, and F1 measures based on the resulting matching. We use \texttt{microsoft/deberta-xlarge-mnli} as the model for the score and report the F1.
\textbf{BLEURT} \cite{sellam2020bleurt} is a learned evaluation metric based on BERT. It is first pre-trained on synthetic examples and then fine-tuned on human judgments for the task of machine translation.

\subsection{LLM-based Evaluation}
We employed GPT-4o in a structured evaluation framework using a carefully designed template with specific criteria and scoring guidelines to ensure consistent and interpretable evaluations across three practical dimensions: Resource Management, Temporal-Location Feasibility, and Individual Impact. Each dimension is scored on a 1-5 scale with justification, assessing aspects like resource utilization, contextual appropriateness, and solution safety that traditional metrics cannot capture. This structured analysis bridges the gap between surface-level metrics and human evaluation by focusing on operational feasibility. We found a strong correlation (r = 0.82) between the
human evaluation (setting \textit{win, tie} and \textit{lose} to (1, 0.5, -1) respectively) and overall judgment. The prompt is available in the appendix.

\subsection{Human Evaluation}
\label{sec:human_eval}
Following \cite{goyal2022news}'s finding that automatic metrics may not reliably evaluate LLM outputs, we conducted human evaluation using 50 evaluators recruited through CloudConnect Research. Evaluators were college-educated residents of English-speaking countries with basic familiarity with disaster scenarios. They assessed 100 randomly sampled question-response pairs for each time frame, comparing LLM outputs against \textsf{RESPONSE} references using a Win/Tie/Lose framework.
Evaluators rated responses on three criteria:
(1) \textbf{Sufficiency}: Whether the answer-explanation pair provides complete, situation-appropriate information
(2) \textbf{Usefulness}: How actionable and valuable the information is for implementation
(3) \textbf{Correctness}: Whether the evaluator would follow the given advice
This evaluation framework helps assess both the quality and practicality of LLM-generated responses in disaster scenarios.

\subsection{Human Performance} For the sake of completeness, we sampled 10 questions for each time frame among the ones evaluated by humans and submit those questions to a pool of 30 human annotators excluding the ones who participated to the data collection and the human evaluation efforts. We collected their answers and evaluated them using the automated evaluation metrics described earlier to confirm the initial findings.

\section{Results}

In this section, we present the numerical results of the experiments detailed earlier and discuss our findings. All the human-evaluated pairs of answer and explanation are outputs from GPT-4. 

\subsection{Automatic Evaluation}
\label{sec:auto_eval}
The automatic evaluation results in \Cref{tab: auto_eval_results} prove inconclusive for three main reasons. First, most metrics show limited sensitivity, with BLEU scores (range [0.0, 100.0]) clustering between 0.42-0.68 and similarly compressed ranges for BLEURT. Second, BertScore values are consistently high across all experiments, contradicting human evaluation results and suggesting it may overemphasize surface-level similarities. Third, the metrics show inconsistent rankings - for instance, in the immediate timeframe, BLEU favors Claude-3 while BertScore favors GPT-4. These inconsistencies suggest that current automatic metrics may be insufficient for evaluating LLM performance on this task, highlighting the need for more sophisticated evaluation approaches.

\subsection{LLM-based Evaluation}
The structured evaluation reveals that human responses consistently outperform LLM solutions across all assessment dimensions, achieving an overall score of 0.81. Among the LLMs, Claude (0.75) performs better than GPT-4 (0.68), though both fall short of human performance. Notably, all sources score highest on Resource Management (Human: 0.83, Claude: 0.77, GPT-4: 0.71), suggesting that identifying and addressing missing resources is handled more effectively than other aspects. The lower scores in Individual Impact (Human: 0.78, Claude: 0.72, GPT-4: 0.64) indicate that providing safe, actionable guidance remains a challenge, particularly for LLMs. These findings align with our human evaluation results, where LLMs showed better performance in longer time horizons but struggled with immediate response scenarios. The results are shown in \Cref{tab:llm_eval}.
\begin{table}[h]
\centering
\resizebox{0.95\columnwidth}{!}{%
\begin{tabular}{@{}lllll@{}}
\toprule
& Resource & Feasibility & Impact & Overall \\ \midrule
Human  & 0.83               & 0.82        & 0.78             & 0.81    \\
GPT-4  & 0.71               & 0.69        & 0.64             & 0.68    \\
Claude & 0.77               & 0.76        & 0.72             & 0.75 \\ \bottomrule
\end{tabular}%
}
\caption{Structured LLM-based evaluation of solutions across three dimensions: resource management, feasibility, and individual impact.}
\label{tab:llm_eval}
\end{table}

\subsection{Human Evaluation}

\begin{table*}[h]
\centering
\resizebox{0.9\textwidth}{!}{%
\begin{tabular}{@{}llllllllll@{}}
\toprule
\multirow{2}{*}{}  & \multicolumn{3}{c}{\textbf{Sufficiency}}       & \multicolumn{3}{c}{\textbf{Usefulness}}       & \multicolumn{3}{c}{\textbf{Correctness}}      \\ \cmidrule(l){2-4} \cmidrule(l){5-7} \cmidrule(l){8-10} 
                   & Win ($\uparrow$) & Tie  & Lose ($\downarrow$) & Win ($\uparrow$) & Tie  & Lose ($\downarrow$) & Win ($\uparrow$) & Tie  & Lose ($\downarrow$) \\ \midrule
\textbf{Immediate} & 31.4             & 9.6  & 59.0                & 37.3             & 5.7  & 57.0                & 37.3             & 5.9  & 56.8                \\
\textbf{Week}      & 56.0             & 2.0 & 42.0                & 56.0             & 4.0 & 40.0                & 58.0             & 2.0  & 40.0                \\
\textbf{Month}     & 62.0             & 8.0 & 30.0                & 64.0             & 8.0 & 32.0                & 57.9             & 12.0 & 30.0                \\ \bottomrule
\end{tabular}}
\caption{Human evaluation. \textit{Sufficiency} penalizes an answer that is not adequate enough to address the problem mentioned in the question. \textit{Usefulness} evaluates the informativeness of the proposed solution and \textit{correctness} measures the likelihood of a human to implement the recommendation described in the solution. The evaluators compared the generated answers against \textsf{RESPONSE} gold references.}
\label{tab:human_eval}
\end{table*}

\begin{table*}[ht]
\centering
\resizebox{0.9\textwidth}{!}{%
\begin{tabular}{@{}llllllllll@{}}
\toprule
\multirow{2}{*}{} & \multicolumn{3}{c}{\textbf{Immediate}}         & \multicolumn{3}{c}{\textbf{Week}}              & \multicolumn{3}{c}{\textbf{Month}}             \\ \cmidrule(l){2-4} \cmidrule(l){5-7} \cmidrule(l){8-10} 
                  & BLEURT         & BLEU          & BertS         & BLEURT         & BLEU          & BertS         & BLEURT         & BLEU          & BertS         \\ \midrule
\textbf{GPT-4}    & \textbf{-0.57} & 1.04          & 0.59          & -0.92          & \textbf{5.27} & \textbf{0.57} & -0.94          & \textbf{1.12} & \textbf{0.55} \\
\textbf{Claude-3}   & -0.82          & 1.02          & 0.55          & -1.15          & 0.68          & 0.55          & -1.06          & 0.79          & 0.54          \\
\textbf{Human}    & -0.79          & \textbf{1.10} & \textbf{0.62} & \textbf{-0.82} & 3.05          & 0.55          & \textbf{-0.84} & 0.97          & 0.54          \\ \bottomrule
\end{tabular}}
\caption{Human performance on the QA task. We evaluate the human answers against \textsf{RESPONSE} gold references and include GPT-4 and Claude-3 results for comparison. Best scores are boldfaced.}
\label{tab:human_perf}
\end{table*}

\paragraph{Are LLMs answers sufficient and adequate?} We evaluated whether LLM responses provide necessary information for taking appropriate action in given scenarios. Following the protocol in \Cref{sec:human_eval}, annotators compared generated answer-explanation pairs with \textsf{RESPONSE} references, marking \textit{Win}, \textit{Tie}, or \textit{Lose}. Results in \Cref{tab:human_eval} (columns 1-3) show that for immediate responses, evaluators preferred LLM outputs in only 31.4\% of cases (59.0\% preferred references, 9.6\% ties). However, LLM performance improved for longer timeframes, with evaluators preferring generated responses in 56\% of week-horizon and 62\% of month-horizon cases. This suggests LLMs are more reliable for longer-term planning than immediate response guidance.

\paragraph{Are the answers of LLMs useful and informative?} This question measures which pair best provides a clear answer with sufficient details to guide someone towards a successful and commonsensical remedial action. We follow the same setting of the previous paragraph. The results are presented in columns 4 to 6 of \Cref{tab:human_eval}. For immediate timeframe pairs, the model often fell short (37.3\% preferred) compared to human-written answers (57\% preferred). However, as the timeframe increases to a week and one month, the model's pair become more helpful (56\% and 64\% preferred respectively). This indicate that the model exhibits a stronger ability to create informative pairs for longer timeframes (weeks and months) compared to immediate ones.

\paragraph{Would you follow this recommendation?} This question is used as a proxy for correctness as it aims to measure how well the generated answers and explanations aligns with human commonsense in incident and disaster situations. This final question follows the same trend of the previous paragraphs. Analyzing \Cref{tab:human_eval}, we find that for immediate pairs (column 7), the model's solutions resonated less with human evaluators, with only a 37.3\% preference. This suggests a need for improvement in remedial recommendation in a context of urgency. However, like in the other cases, there's a positive trend as the timeframe expands to one week and one month (columns 8 and 9). The 'Win' column shows an increase in preference for the model's pairs (58\% and 57.9\% respectively). This indicates a stronger alignment with human judgment for solutions based on longer periods (weeks or months). In summary, the model appears to improve at generating solutions that better match human preferences when asked for remedial actions for incidents spanning longer time horizons.

\subsection{Qualitative analysis}

\Cref{tab:output_imm_part} to \Cref{tab:output_month_part} summarize human evaluations of responses given by GPT-4 
, in comparison to a gold standard reference in \textsf{RESPONSE}. In \Cref{tab:output_imm_part}, GPT-4's first response lacks relevance to the transportation disruption question, focusing on general safety rather than specific solutions. Claude's response is more relevant, directly addressing transportation issues through road safety assessment and alternate routes. In the second scenario, GPT-4's solution is deemed insufficient for the flooding issue. However, in the third example, GPT-4's response aligns well with expectations across all criteria, appropriately suggesting evacuation and authority notification.
Similar patterns emerge in \Cref{tab:output_week_part}. For longer-term scenarios in \Cref{tab:output_month_part}, GPT-4's performance varies: its first response is rated poorly across all criteria (suggesting inappropriate solutions), while its second and third responses score comparably or better than the gold references, particularly in providing actionable recommendations.

\subsection{Human performance}

To establish a human performance baseline, we collected responses from a new set of annotators for a sample of questions and evaluated them using both automatic metrics and human judgment. The results reveal several important insights:
(i) \textit{Metric Limitations}: When evaluating human-generated responses against \textsf{RESPONSE} gold answers using automatic metrics (BLEU, BLEURT, and BertScore), we observed scores within similar ranges as LLM-generated responses. This suggests these metrics may not effectively capture the qualitative differences that matter in disaster response recommendations.
(ii) \textit{Human Agreement}: When we presented both human-generated responses and \textsf{RESPONSE} gold answers to evaluators, approximately 70\% indicated no clear preference between the two. This high rate of "ties" suggests that multiple valid approaches often exist for handling disaster situations, and validates the quality of our dataset's gold answers.
(iii) \textit{Evaluation Consistency}: The structured quality analysis scores (shown in \Cref{tab:llm_eval}) demonstrate a clearer distinction between human and LLM performance than traditional metrics, particularly in Resource Management and Individual Impact dimensions. This suggests our structured analysis framework may better capture meaningful differences in response quality.
These findings highlight the importance of developing more nuanced evaluation methods for disaster response recommendations, where success often depends on practical feasibility rather than textual similarity to a single "correct" answer.

\section{Conclusion}
We release \textsf{RESPONSE}, a question answering dataset for evaluating LLM's reasoning ability in disaster management scenarios. The dataset includes problem identification and time-sensitive solutions, reflecting real-world disaster response complexities. Testing state-of-the-art LLMs on RESPONSE reveals that even GPT-4 achieves only 37\% human-level accuracy for immediate remedial actions. Our results show common evaluation metrics are unsuitable for this task, lacking correlation with human evaluation and disagreeing with each other. This highlights both the need for improved LLM commonsense reasoning in critical decision-making scenarios and the development of better automatic evaluation measures for such tasks.

\section*{Limitations}
Despite the advantages of \textsf{RESPONSE}, we would like to acknowledge certain limitations. Firstly, the dataset inherit inaccuracies of the original Incident1M. Despite efforts to manually select the most expressive images, the labels associated to the images are not always made evident by the image itself. This is mostly reflected in the automatic context generation and in the contextualization of the questions. Second, comparing LLM recommendations with human responses is complex due to the variability in human decision-making. Human responses can vary widely, influenced by personal experiences, expertise, and intuitive thinking, which may not always be consistent or optimal. Hence, aligning machine recommendations with such a varied baseline can be challenging. Third, the subjectivity of the annotations can be a limitation in evaluating LMs as generating a 'correct' answer may depend significantly on individual or cultural perspectives which may not be adequately represented. Additionally, the diversity in types of disasters, cultural contexts, and unforeseen variables present significant challenges that might not be fully addressed.
\section*{Ethical Statement}
For data collection, we relied on annotators located in English speaking regions (Australia, Canada, New Zealand, the United Kingdom, and the United States).  The average hourly pay is 15.00 US dollars, which exceeds the income requirements proposed in the human subjects research protocols. We do not collect personal information about the workers. The interfaces are displayed in the appendix. The proposed dataset does not contain inappropriate suggestions that could potentially exacerbate emergency situations or lead to harm. Due to the experimental nature, limited production applicability, and relatively small dataset scale, we believe the potential for misuse or harm is negligible.
\section*{Dataset Availability}
The \textsf{RESPONSE} dataset and the repository containing the code for generating \textsf{RESPONSE} questions, the evaluation and the generated responses, enabling the reproduction of the results will made publicly available.

\section*{Acknowledgements}

This research was supported by the Leverhulme Trust grant for the project 'Repurposing of Resources: from Everyday Problem Solving through to Crisis Management' (RPG-2021-182). We also thank all reviewers for their insightful feedback.

\bibliography{main}

\appendix

\label{sec:appendix}

\section{Availability and Maintenance}
The \textsc{RESPONSE} dataset and the code repository is made publicly available. The repository contains code for generating \textsc{RESPONSE} questions, the evaluation code and the generated responses, enabling the reproduction of the results presented in this work. We encourage the community to further contribute to develop evaluation metrics using \textsc{RESPONSE}.


\section{Dataset Statistics}

\begin{table}[h]
\centering
\begin{tabular}{@{}ll@{}}
\toprule
&       Value    \\ \midrule
\# of instances        & 1780    \\
\# of sets of questions & 6037    \\
\# of questions        & 18111    \\
Avg/med \# of questions per set & 3.4/3   \\
Avg/med words in problem   & 5.5/5   \\
Avg/med words in $s_1$     & 6.5/6   \\
Avg/med words in $s_2$     & 6.6/6   \\
Avg/med words in $s_3$     & 6.5/6   \\
Avg/med words in $e_1$     & 10.7/10 \\
Avg/med words in $e_2$     & 11.3/11 \\
Avg/med words in $e_3$     & 12.2/11 \\ \bottomrule
\end{tabular}%
    \caption{Dataset statistics. avg: average, med: median, s: solution, e: explanation.}
\label{tab:stats}
\end{table}

\begin{figure}[h]
\centering
\includegraphics[width=0.4\textwidth]{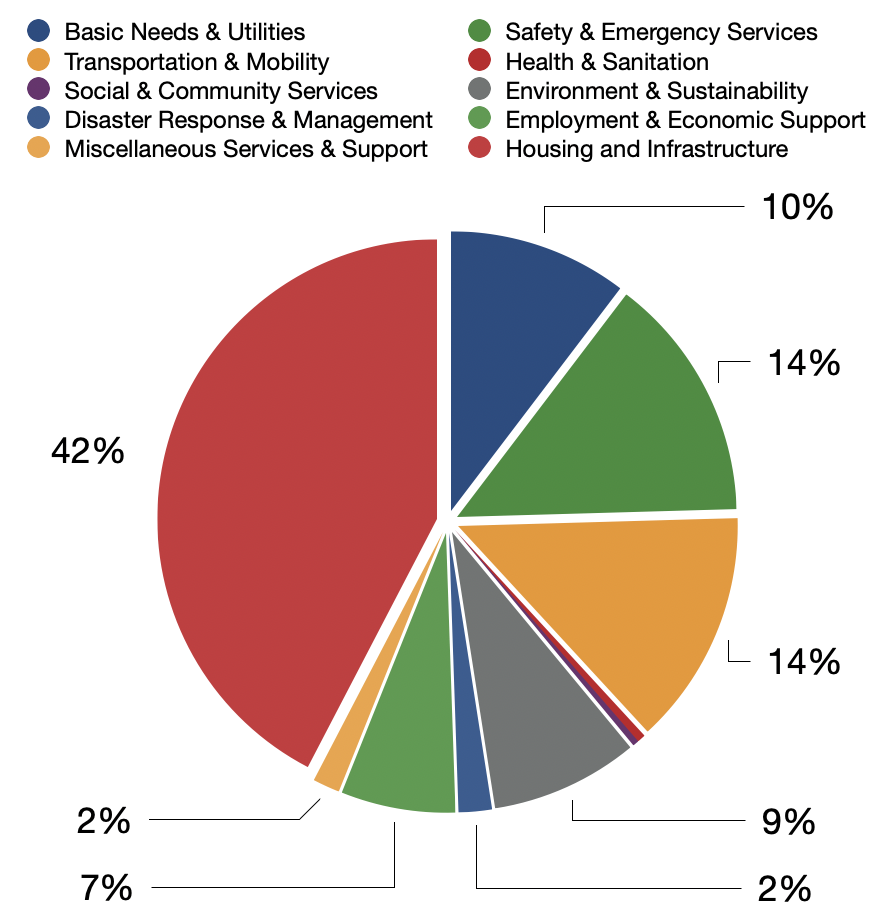}
\caption{Pie chart representing classes for the missing resources in \textsc{RESPONSE}.}
\label{fig:pie_chart}
\end{figure}

\begin{figure}[h]
         \centering
    \includegraphics[width=\columnwidth]{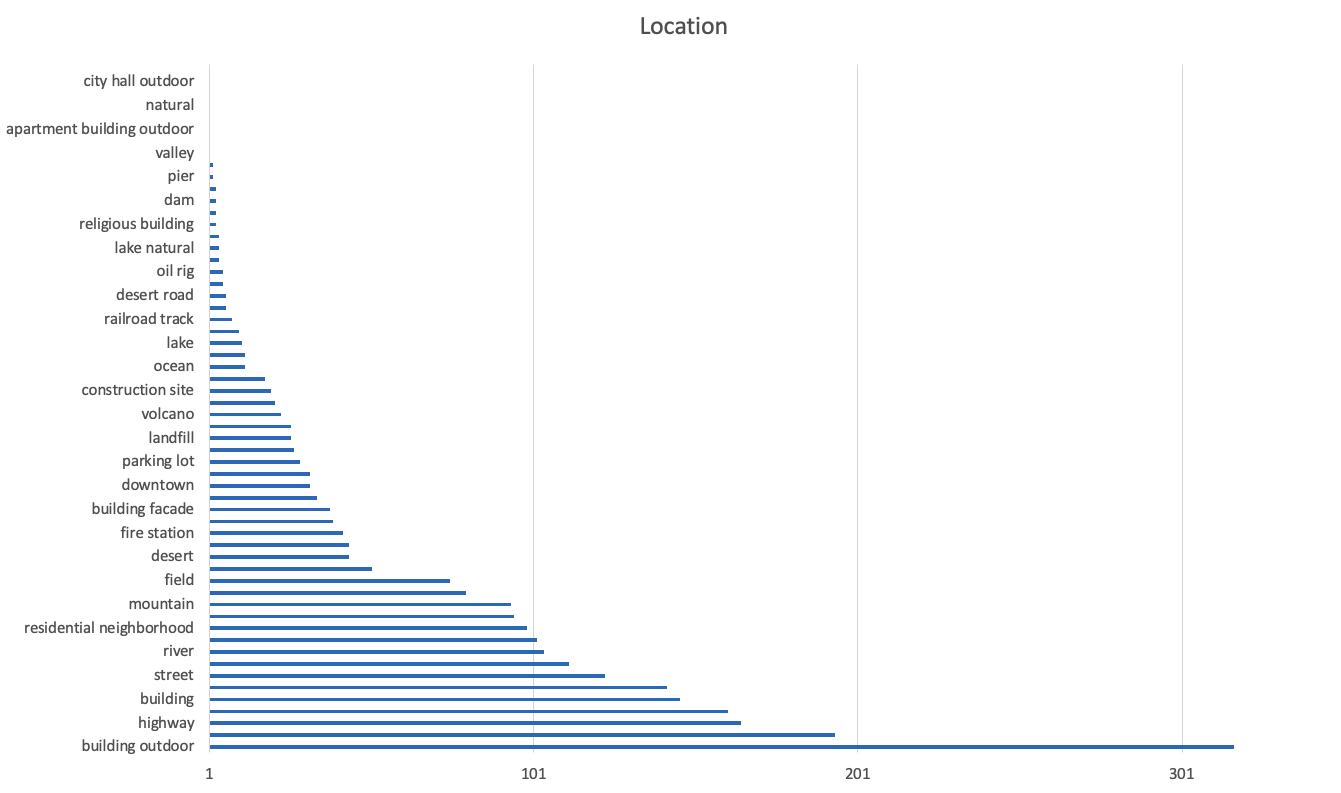}
    \caption{Histograms of location in \textsc{RESPONSE}}
    \label{fig:place_graph}
\end{figure}

\begin{figure}[h]
         \centering
    \includegraphics[width=\columnwidth]{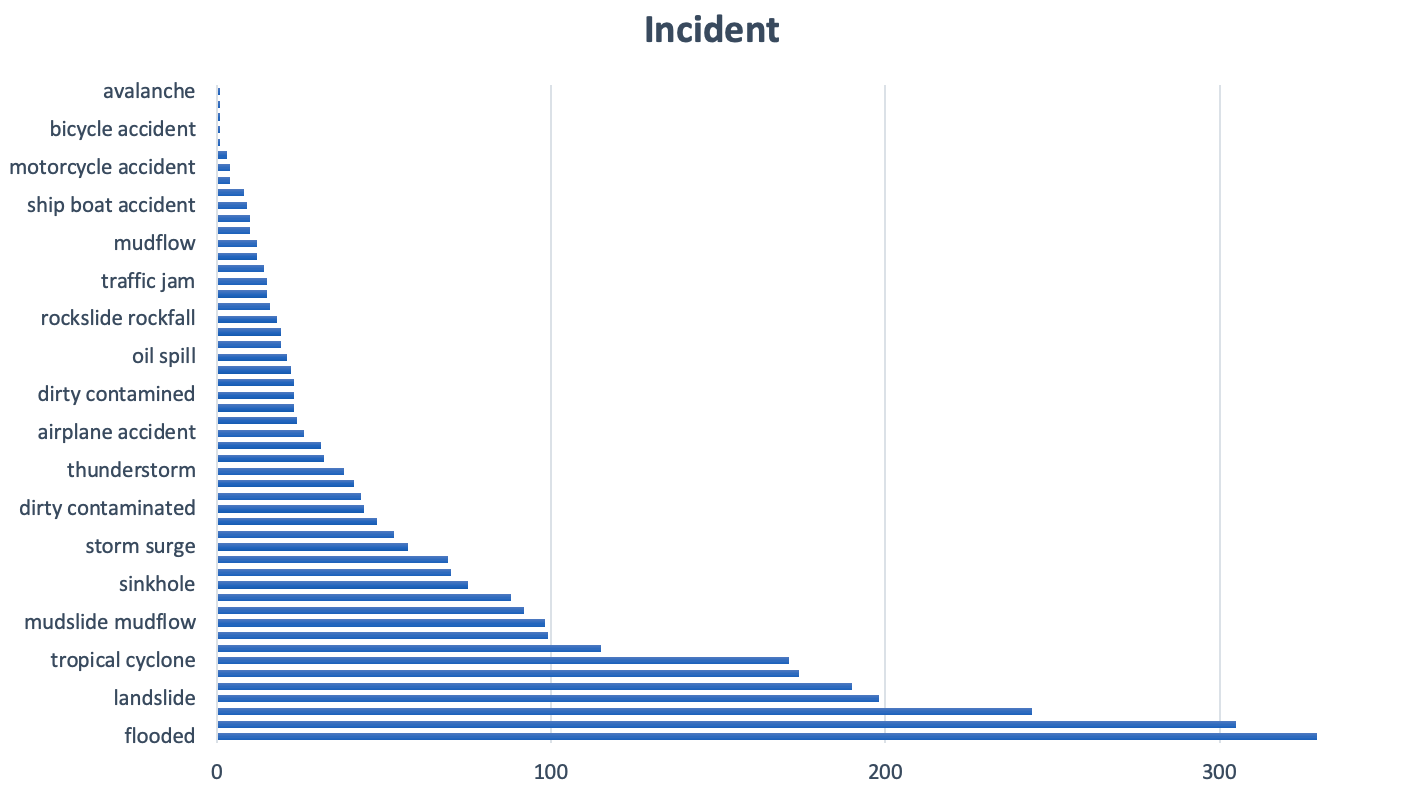}
    \caption{Histograms of incidents in \textsc{RESPONSE}}
    \label{fig:incident_graph}
\end{figure}

\section{Prompts}
\label{sec:prompts}

\subsection{Image captioning}
\textbf{Model}: \texttt{gpt-4o}

\textbf{Parameter}:  temperature set as the default 1.

\textbf{System}: You are a system generating descriptions for incident images. Provided with an image, incident and location if available, you will describe what you see in the image, giving details but staying concise.

\textbf{Content}: What’s in this image?

\subsection{Question generation}
\textbf{Model}: \texttt{gpt-4-turbo}

\textbf{Parameter}:  temperature set to 0.7.

\textbf{System}: You are a helpful assistant that generates questions and outputs your response as pure JSON.
\textbf{Content}: You are given a JSON. 1, 2, and 3 correspond to these timeframes: immediately,  after 1 week, after 1 month. Use the example given for guidance. First, create all the combinations of incident and place values. If place is an empty string, ommit it. Second, create natural, creative and realistic questions concerning the problem, one for each timeframe for which the missing resource is the answer. The question has to ask about remedial actions and responses. Then, paraphrase some of the generated questions (not all of them) (in particular the place if available and timeframes (for example, one month can be transformed in few weeks etc., a week into few days)) to improve the fluidity and the diversity of the questions. Do not mention or hint at the solutions nor the explanations in the generated question. Pay attention to follow the instructions and generate one question per timeframe. Answer in a plain JSON format without any Markdown or code block formatting.

\textbf{Example}: JSON: \{"problem":"getting myself to safety","resource":"ensuring i call 911 and stay far away from the fire","solution1":"move as fast as possible away from the fire","solution2":"get fire fighters to evaluate and stop the fire","solution3":"the fire should be out within 1 month, put preventive measures in place","explanation1":"Life is important, staying away from life reduces the risk to getting injured in the fire","explanation2":"stopping the fire is important","explanation3":"Preventive measures will ensure the fire does not come back", "incident":"wildfire, on fire", "place":"coast, mountain, forest"\}

Answer: \{
  "problem": "getting myself to safety",
  "missing resource": "ensuring I call 911 and stay far away from the fire",
  "scenarios": [
	{
	  "incident": "wildfire",
	  "place": "forest",
	  "question1": "If you encounter a wildfire in a forest, what's the most critical action to take immediately to ensure your safety?",
	  "question2": "After escaping a wildfire in a forest, what's the most important step to take after a week to ensure the fire is contained?",
	  "question3": "A month after a wildfire devastated a forest, what preventative measures can you take to minimize the risk of future fires?"
	},
	{
	  "incident": "on fire",
	  "place": "mountain",
	  "question1": "What's the essential first step to take if you find yourself caught on fire while hiking on a mountain?",
	  "question2": "After escaping a fire situation on a mountain, what's the crucial action to take after a week to ensure the fire is extinguished?",
	  "question3": "One month after a fire ravaged a mountain region, what preventative measures can be implemented to minimize the risk of future fires?"
	},
	{
	  "incident": "wildfire",
	  "place": "coast",
	  "question1": "If you encounter a wildfire near the coast, what immediate action should you take to prioritize your safety?",
	  "question2": "After evacuating from a wildfire on the coast, what's the most important step a week later to ensure the fire is under control?",
	  "question3": "A month after a wildfire swept through a coastal area, what preventative measures can be established to minimize the risk of future wildfires?"
	}
  ]
\}
\subsection{Question Answering}
\textbf{Model}: \texttt{gpt-4-turbo}, \texttt{claude-3-opus-20240229}

\textbf{Parameter}:  temperature set to 0.0.

\textbf{System}: You are a helpful assistant that responds to questions. Imagine you're a human in the situation described by the context. You need to be autonomous in reacting.

\textbf{Content}: "Context: " + \{context\} + "Question: " + \{question\} Respond in maximum 10 words with what you think is the most critical remedial action. Then, explain your answer in maximum 10 words starting with 'Explanation: '

\section{Additional examples}

\begin{table*}[hbt!]
\resizebox{0.95\textwidth}{!}{%
\begin{tabular}{llccc}
\hline
\multicolumn{1}{c}{\textbf{Setting}} & \multicolumn{1}{c}{\textbf{Solution}}                                                                              & \textbf{Suff.}      & \textbf{Usef.}      & \textbf{Correct.}   \\ \hline
\rowcolor{lightgray}
\multicolumn{5}{l}{\textit{A week after a sinkhole event in a building, what steps should be taken to ensure access to food and water?}}                                                                                    \\
 
{Gold}& Answer: look for new routes out of the city.                                                                       & -& -& -\\
{GPT-4}& Answer:secure food, water, shelter, medical supplies.                                                              & -1& -1& -1\\
{Claude}& Answer: relocate affected residents to safe shelter with supplies.                                                 & -& -& -\\ \hline
\rowcolor{lightgray}
\multicolumn{5}{l}{\textit{What are the initial steps to take once it is safe to return to your house after a wildfire?}}                                                                                                   \\
 
{Gold}& Answer:go back to the house and assess damage. start insurance claims.                                             & -& -& -\\
{GPT-4}& Answer: check structural integrity and fire damage.                                                                & -1& 1& -1\\
{Claude}& Answer: check for structural damage and hot spots.                                                                 & -& -& -\\ \hline
\rowcolor{lightgray}
\multicolumn{5}{l}{\textit{What essential action should be taken a few days after a hailstorm to ensure the bridge remains usable?}}                                                                                        \\
 
{Gold}& Answer: hire a snowplow.                                                                                           & -& -& -\\
{GPT-4}& Answer: apply salt and sand regularly.                                                                             & 1& 1& 1\\
{Claude}& Answer: clear snow and ice from bridge surface.                                                                            & -& -& - \\\bottomrule
\end{tabular}}
\caption{Human evaluation for the "week" timeframe. (-1, 0, 1) means (\textit{lose}, \textit{tie}, \textit{win}) respectively.}
\label{tab:output_week_part}
\end{table*}%

\begin{table*}[hbt!]
\centering
\resizebox{0.95\textwidth}{!}{%
\begin{tabular}{@{}lllll@{}}
\toprule
\multicolumn{1}{c}{\textbf{Setting}} & \multicolumn{1}{c}{\textbf{Solution}}                              & \multicolumn{1}{c}{\textbf{Suff.}} & \multicolumn{1}{c}{\textbf{Useful.}} & \multicolumn{1}{c}{\textbf{Correct.}} \\ \midrule
\rowcolor{lightgray}
\multicolumn{5}{l}{\makecell[l]{\textit{A few weeks after experiencing an accident due to an oil spill on the highway, what measures should you take} \\ \textit{for sustained healing?}}}                                                                               \\
 
Gold                                 & Answer: one could look for physical therapy                                & -                                  & -                                  & -                                     \\
GPT-4                                & Answer: counseling for trauma, safety review, infrastructure improvements. & -1                                 & -1                                 & -1                                    \\
 Claude& Answer: counseling& -& -&-\\\midrule
 \rowcolor{lightgray}
\multicolumn{5}{l}{\textit{What long-term actions can be implemented to secure transportation routes from future mudslides along rivers?}}                                                                                  \\
 
Gold                                 &  Answer: study ways to prevent flooding    & -                                  &                                    -&                                       -\\
GPT-4                                & Answer: elevate roads, enforce strict building and zoning regulations.                                     & 0                                  & 0                                  & 1                                     \\
 Claude& Answer: reinforce embankments and improve drainage systems.& -& -&-\\\midrule
 \rowcolor{lightgray}
\multicolumn{5}{l}{\textit{One month after trees have fallen on power lines and blocked a forest, what preventative measures can be set up?}}                                                                               \\
 
Gold                                 & Answer: examine surrounding area for overhead lines near trees             & -                                  &                                    -&                                       -\\
GPT-4                                & Answer: trim and manage vegetation around power lines.                     & 1                                  & 1                                  & 1                                     \\
 Claude& Answer: trim trees near power lines regularly.                                           & -& -&-\\\bottomrule
\end{tabular} }
\caption{Human evaluation for the "month" timeframe. (-1, 0, 1) means (\textit{lose}, \textit{tie}, \textit{win}) respectively.}
\label{tab:output_month_part}
\end{table*}

\begin{table*}[hbt!]
\caption{Sample outputs for the "immediate" timeframe. The scores reported are from the human evaluation. -1 indicate a preference for the gold reference, 1 for the generate answer, 0 no preference.}
\label{tab:output_imm}
\centering
\resizebox{\textwidth}{!}{%
\begin{tabular}{llccc}
\hline
\multicolumn{1}{c}{\textbf{Setting}} & \multicolumn{1}{c}{\textbf{Solution}}                                             & \textbf{Suff.} & \textbf{Usef.} & \textbf{Correct.} \\ \hline
\multicolumn{5}{l}{\makecell[l]{\textit{Immediately after a flood near a river disrupts transportation, what action should be prioritized to}\textit{address the issue?}}}                \\
\multirow{2}{*}{Gold}                & Answer: announce the news of the flooding                                         & \multirow{2}{*}{-}   & \multirow{2}{*}{-}  & \multirow{2}{*}{-}   \\
                                     & Exp: this will keep the residents of the area informed and save lives             &                      &                     &                      \\
\multirow{2}{*}{GPT-4}               & Answer: seek higher ground immediately, avoid flood waters.                       & \multirow{2}{*}{-1}  & \multirow{2}{*}{-1} & \multirow{2}{*}{-1}  \\
                                     & Exp: floodwaters can rise quickly, posing immediate danger.                       &                      &                     &                      \\
\multirow{2}{*}{Claude}              & Answer: assess road safety and establish alternate routes.                        & \multirow{2}{*}{-}   & \multirow{2}{*}{-}  & \multirow{2}{*}{-}   \\
                                     & Exp: ensuring public safety and maintaining essential transportation are crucial. &                      &                     &                      \\ \hline
\multicolumn{5}{l}{\textit{What immediate steps should you take to protect your flat home from becoming flooded near a river?}}                                                              \\
\multirow{2}{*}{Gold}                & Answer: use sand bags to slow down the flow of water.                             & \multirow{2}{*}{-}   & \multirow{2}{*}{-}  & \multirow{2}{*}{-}   \\
                                     & Exp: it's a quick solution and nonpermanent                                       &                      &                     &                      \\
\multirow{2}{*}{GPT-4}               & Answer: avoid contact, report to authorities immediately.                         & \multirow{2}{*}{-1}  & \multirow{2}{*}{0}  & \multirow{2}{*}{-1}  \\
                                     & Exp: quick reporting helps manage and mitigate risks.                             &                      &                     &                      \\
\multirow{2}{*}{Claude}              & Answer: sandbag or barricade entrances and low-lying areas.                       & \multirow{2}{*}{-}   & \multirow{2}{*}{-}  & \multirow{2}{*}{-}   \\
                                     & Exp: prevents water from entering and diverts flow away from home.                &                      &                     &                      \\ \hline
\multicolumn{5}{l}{\textit{What should be done immediately if a forest catches fire to mitigate the impact on life and housing?}}                                                            \\
\multirow{2}{*}{Gold}                & Answer: relocate those with houses nearby to shelters.                            & \multirow{2}{*}{-}   & \multirow{2}{*}{-}  & \multirow{2}{*}{-}   \\
                                     & Exp: this would help save lives in the affected area                              &                      &                     &                      \\
\multirow{2}{*}{GPT-4}               & Answer: evacuate area, alert authorities, assess immediate dangers.               & \multirow{2}{*}{1}   & \multirow{2}{*}{1}  & \multirow{2}{*}{1}   \\
                                     & Exp: prioritize safety, coordinate emergency response efforts.                    &                      &                     &                      \\
\multirow{2}{*}{Claude}              & Answer: evacuate the area to safety. Call firefighters.                           & \multirow{2}{*}{-}   & \multirow{2}{*}{-}  & \multirow{2}{*}{-}   \\
                                     & Exp: prioritize human lives, then property. Alert professionals to contain fire.  &                      &                     &                      \\ \hline
\end{tabular}}
\end{table*}

\begin{table*}[hbt!]
\caption{Sample outputs for the "week" timeframe. The scores reported are from the human evaluation.}
\label{tab:output_week}
\resizebox{\textwidth}{!}{%
\begin{tabular}{llccc}
\hline
\multicolumn{1}{c}{\textbf{Setting}} & \multicolumn{1}{c}{\textbf{Solution}}                                                                              & \textbf{Suff.}      & \textbf{Usef.}      & \textbf{Correct.}   \\ \hline
\multicolumn{5}{l}{\textit{A week after a sinkhole event in a building, what steps should be taken to ensure access to food and water?}}                                                                                    \\
\multirow{2}{*}{Gold}                & Answer: look for new routes out of the city.                                                                       & \multirow{2}{*}{-}  & \multirow{2}{*}{-}  & \multirow{2}{*}{-}  \\
                                     & Exp: you can either leave the area or bring back supplies in that way                                              &                     &                     &                     \\
\multirow{2}{*}{GPT-4}               & Answer:secure food, water, shelter, medical supplies.                                                              & \multirow{2}{*}{-1} & \multirow{2}{*}{-1} & \multirow{2}{*}{-1} \\
                                     & Exp: essentials ensure survival and health post-disaster.                                                          &                     &                     &                     \\
\multirow{2}{*}{Claude}              & Answer: relocate affected residents to safe shelter with supplies.                                                 & \multirow{2}{*}{-}  & \multirow{2}{*}{-}  & \multirow{2}{*}{-}  \\
                                     & Exp: structural damage risks collapse, necessitating evacuation and aid                                            &                     &                     &                     \\ \hline
\multicolumn{5}{l}{\textit{What are the initial steps to take once it is safe to return to your house after a wildfire?}}                                                                                                   \\
\multirow{2}{*}{Gold}                & Answer:go back to the house and assess damage. start insurance claims.                                             & \multirow{2}{*}{-}  & \multirow{2}{*}{-}  & \multirow{2}{*}{-}  \\
                                     & Exp: once the wildfire has died down and it's safe it's important to go back and see if the house can be salvaged. &                     &                     &                     \\
\multirow{2}{*}{GPT-4}               & Answer: check structural integrity and fire damage.                                                                & \multirow{2}{*}{-1} & \multirow{2}{*}{1}  & \multirow{2}{*}{-1} \\
                                     & Exp: ensures safety and stability of buildings.                                                                    &                     &                     &                     \\
\multirow{2}{*}{Claude}              & Answer: check for structural damage and hot spots.                                                                 & \multirow{2}{*}{-}  & \multirow{2}{*}{-}  & \multirow{2}{*}{-}  \\
                                     & Exp: ensure safety before entering and prevent reignition of fire.                                                 &                     &                     &                     \\ \hline
\multicolumn{5}{l}{\textit{What essential action should be taken a few days after a hailstorm to ensure the bridge remains usable?}}                                                                                        \\
\multirow{2}{*}{Gold}                & Answer: hire a snowplow.                                                                                           & \multirow{2}{*}{-}  & \multirow{2}{*}{-}  & \multirow{2}{*}{-}  \\
                                     & Exp: this would ensure the roadway didn't have ice                                                                 &                     &                     &                     \\
\multirow{2}{*}{GPT-4}               & Answer: apply salt and sand regularly.                                                                             & \multirow{2}{*}{1}  & \multirow{2}{*}{1}  & \multirow{2}{*}{1}  \\
                                     & Exp: prevents ice formation, enhances traction.                                                                    &                     &                     &                     \\
\multirow{2}{*}{Claude}              & clear snow and ice from bridge surface.                                                                            & \multirow{2}{*}{-}  & \multirow{2}{*}{-}  & \multirow{2}{*}{-}  \\
                                     & Exp: prevent refreezing, maintain traction, avoid further accidents.                                               &                     &                     &                     \\ \hline
\end{tabular}}
\end{table*}
\begin{table*}[hbt!]
\caption{Sample outputs for the "month" timeframe. }
\label{tab:output_month}
\resizebox{\textwidth}{!}{%
\begin{tabular}{llccc}
\hline
\multicolumn{1}{c}{\textbf{Setting}} & \multicolumn{1}{c}{\textbf{Solution}}                                                    & \textbf{Suff.}      & \textbf{Usef.}      & \textbf{Correct.}   \\ \hline
\multicolumn{5}{l}{\textit{A few weeks after experiencing an oil spill on the highway, what measures should you take for sustained healing?}}                                                     \\
\multirow{2}{*}{Gold}                & Answer: one could look for physical therapy                                              & \multirow{2}{*}{-}  & \multirow{2}{*}{-}  & \multirow{2}{*}{-}  \\
                                     & Exp: long term physical therapy can help with healing better                             &                     &                     &                     \\
\multirow{2}{*}{GPT-4}               & Answer: counseling for trauma, safety review, infrastructure improvements.               & \multirow{2}{*}{-1} & \multirow{2}{*}{-1} & \multirow{2}{*}{-1} \\
                                     & Exp: address psychological, safety, and environmental concerns.                          &                     &                     &                     \\
\multirow{2}{*}{Claude}              & Answer: counseling.                                                                      & \multirow{2}{*}{-}  & \multirow{2}{*}{-}  & \multirow{2}{*}{-}  \\
                                     & Exp: mental health support is crucial after a traumatic event.                           &                     &                     &                     \\ \hline
\multicolumn{5}{l}{\textit{What long-term actions can be implemented to secure transportation routes from future mudslides along rivers?}}                                                        \\
\multirow{2}{*}{Gold}                & Answer: study ways to prevent flooding                                                   & \multirow{2}{*}{-}  & \multirow{2}{*}{-}  & \multirow{2}{*}{-}  \\
                                     & Exp: to make sure this does not happen again take steps to correct the underlying issues &                     &                     &                     \\
\multirow{2}{*}{GPT-4}               & Answer: elevate roads, enforce strict building and zoning regulations.                   & \multirow{2}{*}{0}  & \multirow{2}{*}{0}  & \multirow{2}{*}{1}  \\
                                     & Exp: prevents water from reaching critical infrastructure, enhances safety.              &                     &                     &                     \\
\multirow{2}{*}{Claude}              & Answer: reinforce embankments and improve drainage systems.                              & \multirow{2}{*}{-}  & \multirow{2}{*}{-}  & \multirow{2}{*}{-}  \\
                                     & Exp: strengthening riverbanks and managing water flow prevent mudslides.                 &                     &                     &                     \\ \hline
\multicolumn{5}{l}{\textit{One month after trees have fallen on power lines and blocked a forest, what preventative measures can be set up?}}                                                     \\
\multirow{2}{*}{Gold}                & Answer: examine surrounding area for overhead lines near trees                           & \multirow{2}{*}{-}  & \multirow{2}{*}{-}  & \multirow{2}{*}{-}  \\
                                     & Exp: this will secure the lines to clear trees away from lines proactively               &                     &                     &                     \\
\multirow{2}{*}{GPT-4}               & Answer: trim and manage vegetation around power lines.                                   & \multirow{2}{*}{1}  & \multirow{2}{*}{1}  & \multirow{2}{*}{1}  \\
                                     & Exp: prevents obstructions and enhances line safety.                                     &                     &                     &                     \\
\multirow{2}{*}{Claude}              & Answer: trim trees near power lines regularly.                                           & \multirow{2}{*}{-}  & \multirow{2}{*}{-}  & \multirow{2}{*}{-}  \\
                                     & Exp: proactive trimming reduces risk of fallen trees on lines.                           &                     &                     &                     \\ \hline
\end{tabular}}
\end{table*}

\clearpage
\onecolumn
\section{Data Collection} 

\begin{figure*}[hbt!]
    \centering
    \includegraphics[width=0.92\textwidth]{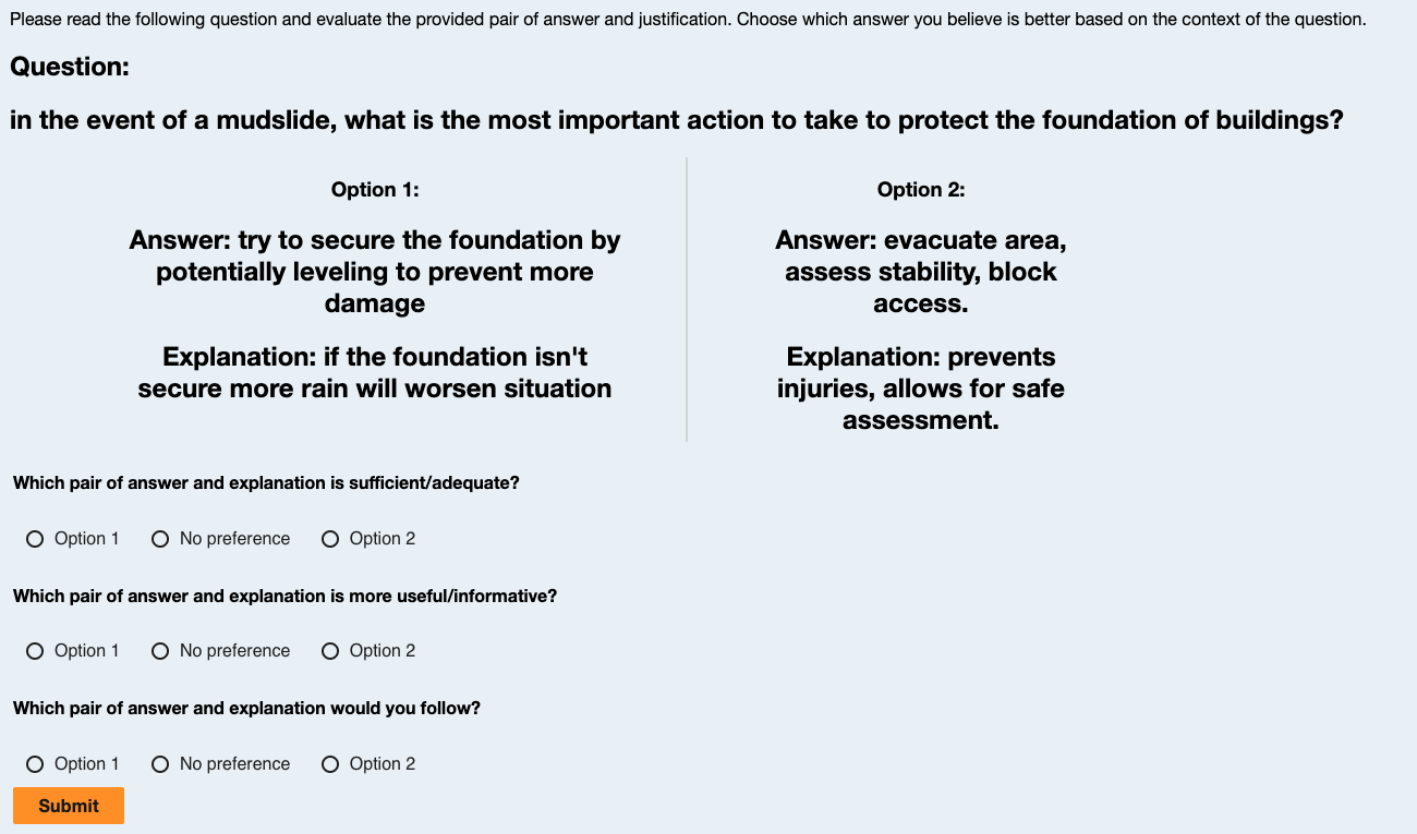}
    \caption{CloudConnect Research Interface for Data Evaluation}
    \label{fig:enter-label}

    \centering
    \includegraphics[width=0.92\textwidth]{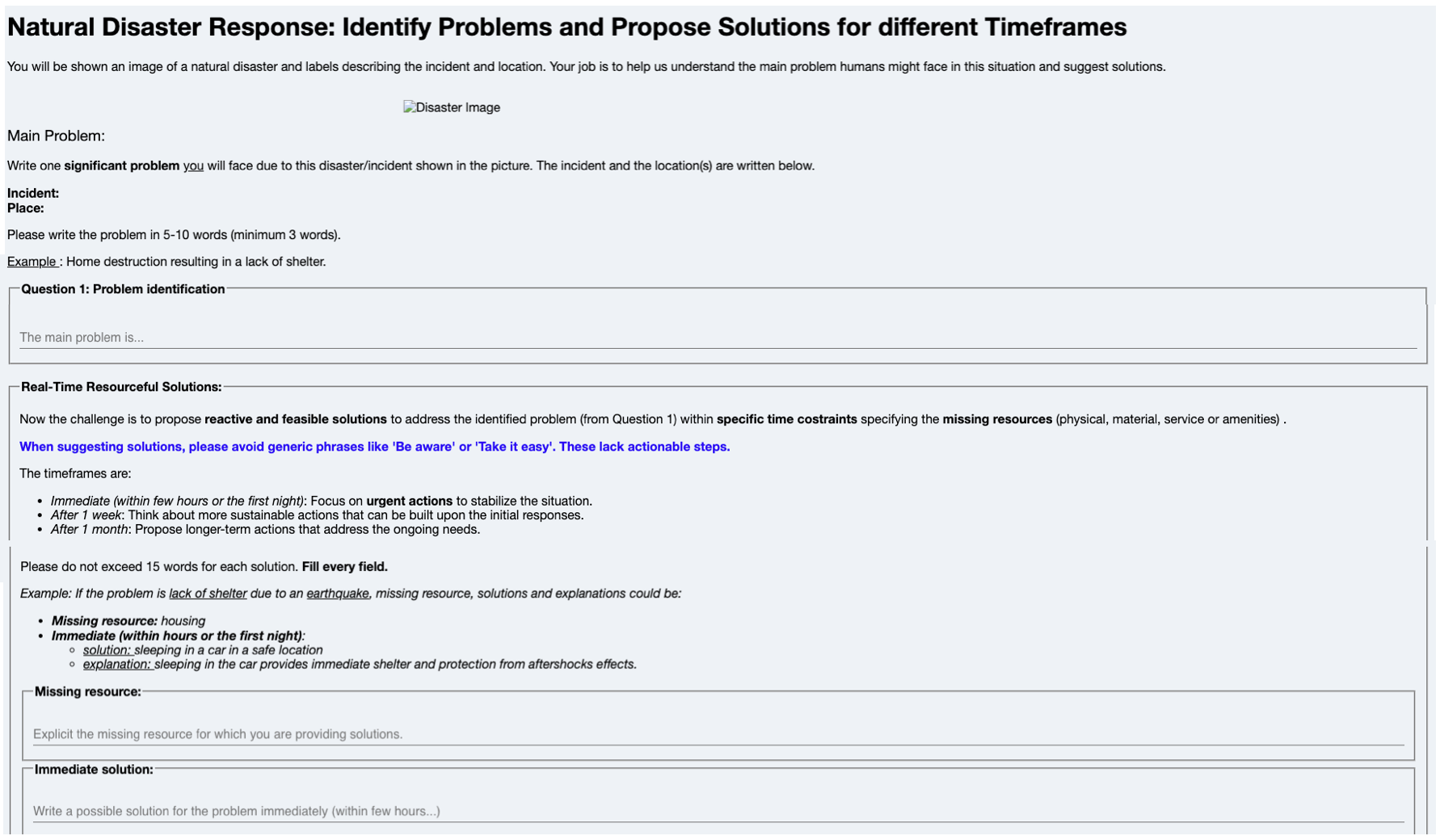}
    \caption{CloudConnect Research Interface for Data Collection}
    \label{fig:enter-label}
\end{figure*}


\end{document}